\pdfoutput=1

\documentclass[11pt]{article}

\usepackage{acl}

\usepackage{times}
\usepackage{latexsym}
\usepackage{amssymb,graphicx}
\usepackage{epsfig}

\usepackage[T1]{fontenc}

\usepackage[utf8]{inputenc}

\usepackage{microtype}

\usepackage{xcolor}
\usepackage[normalem]{ulem}
\usepackage{graphicx}
\usepackage{footnote}
\makesavenoteenv{tabular}
\makesavenoteenv{table}
\usepackage{multicol}
\usepackage{soul}
\usepackage{enumitem}
\usepackage{multicol,caption}

\usepackage{comment}

\usepackage{color, colortbl}

\usepackage{booktabs}
\usepackage{setspace}

\usepackage[textsize=footnotesize]{todonotes}

\newcommand{\bs}[1]{{\textcolor{teal}{#1}}{\bf}}

\begin{document}


\title{
%
%
{Under the {M}orphosyntactic {L}ens:\\
{A} {M}ultifaceted {E}valuation of {G}ender {B}ias in {S}peech {T}ranslation  }
}

\author{Beatrice Savoldi\textsuperscript{1,2}, Marco Gaido\textsuperscript{1,2}, Luisa Bentivogli\textsuperscript{2}, Matteo Negri\textsuperscript{2}, Marco Turchi\textsuperscript{2} \\
  \textsuperscript{1} University of Trento \\
  \textsuperscript{2} Fondazione Bruno Kessler \\
   \texttt{beatrice.savoldi@unitn.it}\\
  \texttt{\{mgaido,bentivo,negri,turchi\}@fbk.eu}}
  


\date{}

\maketitle
\begin{abstract}


Gender bias is 
largely recognized as a problematic phenomenon affecting language technologies, with recent studies underscoring that  
it might surface differently across languages. 
However, 
most of current evaluation practices 
adopt a 
word-level focus on a 
narrow
set of occupational 
nouns
under
synthetic conditions. 
Such protocols overlook
key
features 
of grammatical gender languages, which are characterized by morphosyntactic 
chains of gender agreement, marked on a variety of lexical items and parts-of-speech (POS). 
%
%
%
%
%
%
%
%
To overcome this limitation, we enrich the natural, gender-sensitive MuST-SHE corpus \cite{bentivogli-2020-gender} with two new linguistic annotation layers   
(POS and agreement chains), and
explore to what extent different
lexical categories and agreement phenomena are impacted by gender skews. 
Focusing on speech translation,
we conduct 
a multifaceted evaluation
on three language directions (English-French/Italian/Spanish), with 
models trained on varying amounts of data and different word segmentation techniques.
%
By shedding 
light on model behaviours, gender bias, and its detection at several levels of granularity, 
our findings emphasize the 
value
of dedicated 
analyses beyond aggregated overall results.




\end{abstract}

\section{Introduction}
As \citet{matasovic2004gender} posits: ``\textit{Gender is perhaps the only grammatical category that ever evoked passion -- and not only among linguists.}'' 
That is because, in the case of human entities, 
masculine or feminine inflections are assigned semantically, i.e. in relation to the extra-linguistic reality of gender \cite{ackerman2019syntactic, Corbett:91, corbett2013expression}.
%
%
Thus,
gendered features 
interact with the 
-- sociocultural and political -- 
perception and representation of individuals \cite{Gygaxindex4},
%
by prompting discussions 
on the appropriate recognition 
of gender groups
and their linguistic visibility \cite{stahlberg2007representation, bookgender, hord2016bucking}. 
Such concerns 
also invested 
language technologies \cite{sun-etal-2019-mitigating, cao-daume-iii-2020-toward}, 
where it has been shown that automatic translation systems tend to over-represent masculine forms 
and amplify stereotypes 
when translating 
into grammatical gender languages \cite{savoldi2021gender}. 


Current evaluation practices for assessing gender bias in both Machine (MT) and Speech Translation (ST)
commonly inspect such concerning behaviours 
by focusing only on a restricted set of occupational nouns 
(e.g. \textit{nurse}, \textit{doctor}), and on synthetic benchmarks
\cite{stanovsky-etal-2019-evaluating, escude-font-costa-jussa-2019-equalizing, renduchintala-etal-2021-gender}. Also, {even} when relying on lexically richer natural benchmarks, the designed metrics still work at the word level, treating all gender-marked words indiscriminately \cite{alhafni-etal-2020-gender, bentivogli-2020-gender}. 
Accordingly, current test sets and protocols: \textit{i)} do not allow us to inspect if and to what extent different word categories 
participate
in gender bias, \textit{ii)} 
overlook the
underlying morphosyntactic nature of grammatical gender on agreement chains, which cannot be monitored on single isolated words 
(e.g. \textit{en}: a strange friend; \textit{it}: un\underline{a/o} stran\underline{a/o} amic\underline{a/o}).
In fact, to be grammatically correct, each word in the chain has to be inflected with the same (masculine or feminine) gender form.\footnote{For an analogy, consider the case of (lack of) number agreement in the following: ``*a dogs barks''.}


We believe that fine-grained evaluations including the analysis of 
gender agreement across different parts of speech (POS)
are relevant not only to gain a deeper understanding of bias in
grammatical  
gender languages, but also to inform 
mitigating strategies and data curation procedures.

Toward these goals, our contributions are as follows. \textbf{(1)}
We enrich MuST-SHE \cite{bentivogli-2020-gender} -- the only natural
gender-sensitive benchmark available for 
MT and also
ST -- with two
layers of linguistic 
information:
POS and agreement chains.\footnote{The annotation layers are an extension of  MuST-SHE v1.2 and  are freely downloadable at: \url{ict.fbk.eu/must-she/} under the same MuST-SHE licence (CC BY NC ND 4.0)}
\textbf{(2)}
In light of recent studies 
exploring how 
 model
 design and overall 
 perfomance interplay with gender bias \cite{ roberts2020decoding, gaido-etal-2021-split}, we rely on 
our 
manually curated resource
to compare 
three
ST models, which 
are trained on varying amounts of data, and built 
with different segmentation 
techniques: character and byte-pair-encoding (BPE) \cite{sennrich-etal-2016-neural}.
%

We carry out a multifaceted evaluation that includes automatic and extensive manual analyses on three language pairs (en-es, en-fr, en-it) and  
we consistently find that: \textit{i)}
not all  
POS 
are 
equally impacted by gender bias;
\textit{ii)} 
translating words in agreement
does not emerge as a systematic 
issue; \textit{
iii)} ST systems produce a considerable amount of 
neutral rewordings in lieu of gender-marked expressions, which current binary benchmarks fail to recognize. Finally, in line with concurring
studies, we find that \textit{iv)} 
character-based systems 
have an edge on translating gender phenomena, 
by favouring morphological and lexical diversity.

\section{Background}
\label{sec:background}

While research in Natural Language Processing (NLP) initially 
prioritized narrow technical interventions to address the social impact of language technologies, we are recently attesting a shift toward a more 
comprehensive
understanding of bias \cite{shah-etal-2020-predictive, blodgett-etal-2020-language}. 
%
%
Along this line, 
focus has been 
given 
to bias analysis in 
models'
innards and outputs
\cite{NEURIPS2020_92650b2e, costajussa2020gender}, and to ascertain the validity of bias measurement practices 
\cite{blodgett-etal-2021-stereotyping, antoniak-mimno-2021-bad,goldfarb-tarrant-etal-2021-intrinsic}.
%
Complementary 
mounting
evidence suggests that -- rather than striving for generalizations --
gender bias detection 
ought to incorporate 
contextual and linguistic specificity \cite{gonzalez2020type, ciora2021examining, matthews2021gender, malik2021socially,kurpicz2021world},  
which however receives
little attention due to a heavy focus on English NLP \bs{\cite{bender-friedman-2018-data}.} 
%
Purported agnostic approaches and evaluations \cite{bender-2009-linguistically} can  
prevent from drawing reliable conclusions and mitigating recommendations, 
%
%
%
%
%
%
as attested by monolingual studies on grammatical gender languages 
\cite{zhou2019examining, gonen2019does, zmigrod-etal-2019-counterfactual} and 
in automatic translation scenarios
\cite{vanmassenhove-etal-2018-getting, moryossef-etal-2019-filling}.
%
%
%
%
%
%
%
\begin{figure}[t]
    \centering
    \includegraphics[width=\linewidth]{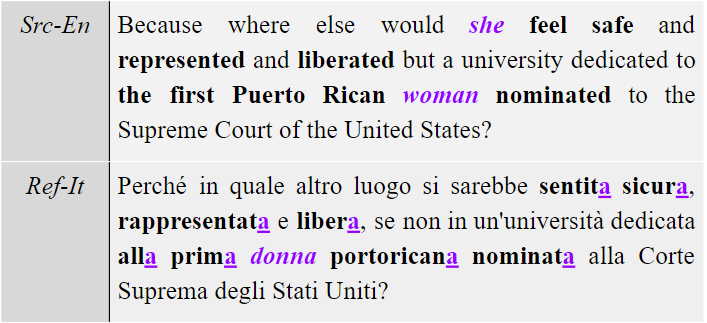}
    \caption{
    Example of gender-mapping in translation from the parallel en-it portion of the natural MuST-SHE corpus. 
    Unlike English, where gender is only expressed on few lexical and pronominal items (\textit{she}, \textit{woman}), in  a grammatical gender language like Italian, 
    gender inflections (here feminine -\underline{a}) are expressed
    on several linguistic items (e.g. verb-\textit{sentita}, adjective-\textit{sicura}) that are in agreement.}
    \label{fig:eg.MuST-SHE}
\end{figure}

Unlike English, 
grammatical gender languages exhibit an elaborate morphological and syntactic system, 
where
gender is overtly marked on numerous POS (e.g., verbs, determiners, nouns), and related words have to agree on the same gender features (see Figure \ref{fig:eg.MuST-SHE} for an example). 
Still, current corpora and evaluation practices do not fully foreground systems’ behaviour on such grammatical constraints.


WinoMT \cite{stanovsky-etal-2019-evaluating} 
represents the 
standard corpus to evaluate gender bias in MT within an English-to-grammatical gender language scenario. It has
been progressively enriched with new features \cite{saunders-etal-2020-neural, kocmi2020gender}, and adapted for ST \cite{costa2020evaluating}. While 
%
this resource can be useful to diagnose
gender stereotyping at scale, 
it 
excludes languages' peculiarities 
since 
it is built on the concatenation of 
two 
corpora designed for English monolingual tasks\footnote{\citet{gonzalez2020type} note that the 
U.S. labor market statistics employed to define stereotypical associations are not always in line with other national gender statistics, thus they may impose an Anglo-centric frame for the detection of bias in other language scenarios.}
-- WinoGender \cite{rudinger2018gender} and WinoBias \cite{zhao-etal-2018-gender} -- which 
consist of synthetic sentences with the same structure
and a pre-selected occupational lexicon (e.g. ``The lawyer yelled
at the \textbf{hairdresser} because \textit{he} did a bad job'').\footnote{\citet{levy2021collecting} recently created BUG on natural English data, but still it is limited to the evaluation of occupations. }
To increase 
variability, \citet{troles2021extending} extend WinoBias by accompanying occupations with highly gender-stereotypical verbs and adjectives. Their evaluation though, still only considers the translated professions as to verify 
if
the 
co-occuring words might skew the models' assumptions. 
%
%
However, gender-marking involves also several other, so far less
accounted POS categories, but 
if
they are just as problematic is not clear yet.

Existing bilingual \cite{alhafni2021arabic}, and multilingual \cite{bentivogli-2020-gender} natural benchmarks, instead, are manually curated 
as 
to identify a
variety of gender phenomena specifically modeled on the accounted languages. 
As a result, they maximize 
lexical and contextual
variability to inspect 
whether 
translation models yield feminine under-representation in real-world-like scenarios \cite{savoldi2021gender}.
However, since this variability is not mapped into 
fine-grained linguistic information, evaluations on such corpora do not
single out
which
instances may be more responsible for gender 
bias.
%
Finally, by
considering each word in isolation, 
they neglect the underlying 
features of gender agreement, which 
determine the grammatical acceptability of 
the 
translation.
%

To the best of our knowledge, only two works
have currently interplayed issues of syntactic agreement and gender bias. 
\citet{renduchintala2021investigating} designed a set of English 
sentences 
involving
a syntactic construction that requires 
to translate 
an occupational term 
according to
its unequivocal ``gender trigger''
(e.g. that \textit{nurse} is a funny \underline{man}). While they find that MT struggles even 
in
such a simple 
setting,
they 
only inspect the translation of a single disambiguated word (\textit{nurse}) rather than 
 a whole group of words in agreement. 
Closer to our intent, \citet{gaido-etal-2020-breeding} analyze the output of different ST systems and note that 
their 
models seem to wrongly pick 
divergent
gender inflections for unrelated words 
in
the same sentence 
(e.g. \textit{en}: As a researcher, professor; \textit{fr}: En tant que cherch\underline{euse}$_{F}$, profess\underline{eur}$_{M}$) but not for dependency-related ones (e.g. \textit{en}: The classic Asian student; \textit{it}: [L\underline{a} classic\underline{a} studente\underline{ssa} asiatic\underline{a}]$_{F}$). 
%
Although limited in scope, their observation is worth being explored systematically.
We thus conduct the very first study that intersects POS, agreement, and gender bias. 
%


\section{MuST-SHE Enrichment}
In light of the above, a fine-grained evaluation of 
bias focused on POS and
gender agreement  requires the creation of a new dedicated resource.
Rather than building it from scratch, we 
%
%
add two annotation layers to the existing MuST-SHE benchmark \cite{bentivogli-2020-gender}, which is built on spoken language data retrieved from TED talks.
%
%
%
%
Available for en-es/fr/it,
it 
represents the only multilingual 
MT and ST \textit{GBET}\footnote{\textit{Gender Bias Evaluation Testset} \citep{sun-etal-2019-mitigating}.} 
exhibiting
a natural variety of 
gender 
phenomena, which are balanced across feminine and masculine forms.

In 
the 
reference translations of the corpus, each target gender-marked word -- 
corresponding 
to a
neutral expression in the English source
--  is annotated 
with its alternative \textit{wrong} gender form 
(e.g. \textit{en}: \textit{\textbf{the}
girl \textbf{left}}; \textit{it}: \textit{\textbf{la<il>} ragazza \`e \textbf{andata<andato>} via}).
As further discussed in in Section \ref{subsec:evaluation}, such a feature enables fine-grained analyses of gender realization, which can also disentangle systems' tendency to (over)generate  masculine -- over feminine forms -- in translation.
%


MuST-SHE thus allows the identification and pinpointed evaluation of numerous and qualitatively different grammatical gender instances under authentic conditions. Furthermore, the target languages covered in MuST-SHE (es, fr, it) are particularly suitable to focus on linguistic specificity. 
As a matter of fact, as \citet{Gygaxindex4} suggest, accounting for 
gender in languages with similar 
typological features allows for proper comparisons.%
\footnote{We underscore that our dedicated resources and experiments intentionally account for the specificities of 
three
(comparable) grammatical gender languages. Hence, we 
remain cautious of
extending by default the results of our annotation and experiments
to any other language.}

%

%
%


\begin{table}[!]
\centering
\setlength{\tabcolsep}{3pt}
\scriptsize
\begin{tabular}{lll}
  \toprule
& &   \textbf{PARTS-OF-SPEECH}  \\
 \cmidrule{1-3}
(a) & \textsc{src} &  As \textit{one} of the \textit{first} women... 
\\ 
 & \textsc{Ref}$_{fr}$ & En tant que l'\textbf{une}$_{Pron}$ des \textbf{premières}$_{Adj-det}$ femmes..\\
  \cmidrule{2-3}
 (b) & \textsc{src} & 
As a \textit{child} \textit{growing up} in Nigeria...  \\ 
 & \textsc{Ref}$_{it}$ & Da \textbf{bambino}$_{Noun}$ \textbf{cresciuto}$_{Verb}$ in Nigeria.\\
 \cmidrule{2-3}
 (c) &\textsc{src} & Then \textit{an amazing} colleague...\\
 & \textsc{Ref}$_{es}$  & Luego \textbf{una}$_{Art}$ \textbf{asombrosa}$_{Adj-des}$ colega...\\ 
 \midrule
&  &   \textbf{AGREEMENT} \\
   \midrule
   (d) & \textsc{src} & I was \textit{the first Muslim} homecoming queen,
   \\
& & \textit{ the first} Somali student \textit{senator}... \\
& \textsc{Ref}$_{es}$ & Fui [\textbf{\underline{la primera}} reina \textbf{\underline{musulmana}}] del baile, 
\\  
 & & [\textbf{\underline{la primera senadora}}] somalí estudiantil... \\ 
 \cmidrule{2-3}
 (e) & \textsc{src} & She's also \textit{been interested} in 
 research. \\ 
& \textsc{Ref}$_{it}$ & E' [\textbf{\underline{stata}} anche \textbf{\underline{attratta}}] dalla ricerca 
.\\
\cmidrule{2-3}
 (f) & \textsc{src} & I also \textit{became} \textit{a} high school \textit{teacher}.  \\ 
& \textsc{Ref}$_{fr}$ & Je suis aussi [\textbf{\underline{devenu un professeur}}] de lycée.  
\\
 \bottomrule
\end{tabular}
\caption{MuST-SHE target \textbf{gender-marked words }annotated per $_{POS}$ and [\underline{agreement chains}]. For the sake of simplicity, the alternative <\textit{wrong} gender-marked words> are not shown.}
\label{tab:annotation_examples}
\end{table}

\subsection{Phenomena Categorization}
\label{subsec:categorization}


\paragraph{Parts-Of-Speech.}
We annotate each target gender-marked word in MuST-SHE with POS information.
%
 %
As shown in Table \ref{tab:annotation_examples} (\textit{a-c}), 
we differentiate among 
six
POS categories:\footnote{
Some POS categories (e.g. conjunctions, adverbs) are not considered since they are not subject to gender inflection.
} 
%
%
\textit{\textbf{i)}} articles,  
\textit{\textbf{ii)}} pronouns, \textit{\textbf{iii})} nouns, and \textit{\textbf{iv})} verbs. 
%
For adjectives, we further distinguish 
\textit{\textbf{v)}} limiting adjectives with minor semantic import  
that determine e.g. possession, quantity, space  (\textit{my}, \textit{some}, \textit{this}); and \textit{\textbf{vi)}} descriptive adjectives 
that convey 
attributes and qualities, e.g. \textit{glad}, \textit{exhausted}.
This distinction enables to neatly sort  our POS 
categories into
the closed 
class of function words, or into the open 
one
of content words \cite{shopen2007parts}. Since 
words from these two classes 
differ substantially in terms of variability, frequency, and 
semantics, 
we reckon they represent a relevant variable to account for in the evaluation of gender bias. 


 \paragraph{Agreement.}
 
 
 
We also enrich MuST-SHE with linguistic information that is relevant to investigate the morphosyntactic nature of grammatical gender agreement.  
%
Gender agreement, 
or
\textit{concord} \citep{corbett2006agreement, comrie1999grammatical},
requires 
that related words match the same gender form,
as in the case of \textit{phrases}, i.e. groups of words that constitute a single linguistic unit.\footnote{If agreement is not respected, the 
 unit becomes ungrammatical e.g. \textit{es}: 
 *el$_{M}$ buen${_M}$ nin\~a$_{F}$
 (\textit{en:} the good kid).} 
%
%
%
%
%
 %
 %
 Thus, as shown in Table \ref{tab:annotation_examples}, we identify and annotate as agreement chains
 gender-marked words that constitute a phrase,  such as a noun plus its modifiers (\textit{d}), and verb phrases for compound tenses (\textit{e}). 
  Also, structures that involve a gender-marked (semi-) copula verb and its predicative complement are annotated as chains (\textit{f}), although in such cases the agreement
constraint is ``weaker''.\footnote{Such structure, due to the  semantics of some linking verbs,  can enable more flexibility.  E.g. in French, \textit{Elle est devenue$_{F}$ un$_{M}$ canard$_{M}$} (\textit{She became a duck}) is grammatical, although \textit{un canard} (a duck) is formally masculine.}
This annotation 
lets
us verify whether a model 
consistently 
picks
the same gender paradigm 
for all words in the chain, 
enabling the assessment of its syntagmatic behaviour.

\subsection{Manual annotation}
POS and 
agreement annotation was manually carried out 
by 6 annotators (2 per language pair) 
undergoing a linguistics/translation studies MA degree,
and with  
native/excellent proficiency in
the assigned target language. 
For each language pair, 
they
annotated the whole corpus independently,
based on detailed guidelines 
(see Appendix \ref{app:annotations}).
%
%
%
For POS, 
we computed inter-annotator agreement (IAA) on
label assignment with the kappa coefficient (in Scott’s $\pi$ formulation) \citep{scott1955reliability}. The resulting values of 0.92 (en-es), 0.94 (en-fr) and 0.96 (en-it) correspond to ``almost perfect'' agreement according to its standard interpretation \cite{landis}.
%
%
For gender agreement,
IAA
was calculated 
on the exact match of the 
complete chains in the two annotations. The resulting Dice coefficients \cite{dice1945measures} 
of 89.23\% (en-es), 93.0\% (en-fr), and 94.34\% (en-it),
can be considered  highly satisfactory given the more complex nature of this latter task. Except for few 
liminal 
cases that were excluded from the dataset, all disagreements were reconciled.

We show the final annotation statistics 
in Table \ref{tab:annotation}. 
Variations 
across languages
are due to inherently cross-lingual differences.\footnote{Spanish, for instance, relies less than French or Italian on the gender-enforcing \textit{\underline{to be}} auxiliary, 
resulting in less gender-marked 
verbs (\textit{fr}: \underline{est} parti/ie; \textit{it}: \underline{\`{e}} partita/o;  \textit{es}: se ha ido).}
 While their discussion is beyond the scope of this work, 
%
overall these figures underscore the so far largely unaccounted 
variability of gender across 
lexical
categories.

\begin{table}[t]
\scriptsize
\centering
\begin{tabular}{l||ccc|c}
\hline
           &       \textbf{en-es} & \textbf{en-fr} & \textbf{en-it} & \textbf{M-SHE All} \\
                    \hline \hline
%
\textbf{POS}            (tot)     & 2099           & 1906           & 2026           & 6031 \\       
\hspace{0.4cm}\textit{Art}        & 487            & 325            & 413           & 1225               \\
\hspace{0.4cm}\textit{Pronoun}    & 104             & 61            & 48            & 213                \\
\hspace{0.4cm}\textit{Adj-det}    & 118            & 106            & 149           & 373                \\
\hspace{0.4cm}\textit{Adj-des}    & 676            & 576            & 448           & 1700               \\
\hspace{0.4cm}\textit{Noun}       & 607            & 344            & 346           & 1297               \\
\hspace{0.4cm}\textit{Verb}       & 107            & 494            & 622           & 1223  \\
\hline
\textbf{AGR-CHAINS} & 420 & 293 & 421 & 1080 \\
\end{tabular}
\caption{Distribution of 
POS and agreement chains per each language and in the whole MuST-SHE corpus.}
\label{tab:annotation}
\end{table}

%


\section{Experimental Setting}
\label{sec:experimental_setting}

\subsection{Speech Translation models}
Our experiments draw on 
studies exploring 
the relation between 
overall 
system
performance, 
model size
and gender bias.
%
%
%
\citet{NEURIPS2020_92650b2e} posit that 
bias increases with model size as larger systems 
better emulate
biased training data.
Working 
on WinoMT/ST, \citet{kocmi2020gender} correlate higher BLEU scores and 
gender stereotyping,
whereas \citet{costa2020evaluating} show that 
systems with lower performance tend to produce fewer feminine translations for 
occupations,
but rely less on stereotypical cues. 
To account for these findings and inspect 
the behavior of  different 
models
under natural conditions, 
we experiment 
with three end-to-end ST 
solutions, namely:
\textsc{large-bpe}, \textsc{small-bpe}, and \textsc{small-char} (see Appendix \ref{app:models} for complete details about the models and training setups).

Developed to achieve state-of-the-art performance, 
\textbf{\textsc{large-bpe}} models
rely on Transformer \citep{vaswani2017attention} and are trained in rich data conditions (1.25M ASR/ST utterances)
by applying BPE segmentation
\cite{sennrich-etal-2016-neural}. 
To achieve high performance, 
we made use of: \textit{i)} 
all the available ST training corpora
for the languages 
addressed, namely 
MuST-C \cite{MuST-Cjournal} and Europarl-ST \citep{europarlst};
\textit{ii)} 
consolidated data augmentation 
methods \citep{nguyen2019improving, Park_2019, jia2018leveraging};
and \textit{iii)} 
knowledge transfer techniques from ASR and 
MT, 
namely component pre-training and knowledge distillation \citep{weiss17, bansal-etal-2019-pre}.\footnote{We are aware 
that  
both
MuST-C
and Europarl-ST are 
characterized by a majority (70\%) of masculine speakers \cite{gaido-etal-2020-breeding, vanmassenhove-etal-2018-getting}.
Although comprehensive statistics are not available for the other ASR and MT training resources, we can reasonably assume they are similarly biased.}
%
%
%
%
%
%
%
%
In terms of 
BLEU score -- 34.12 on en-es, 40.3 on en-fr, 27.7 on en-it --
our \textsc{large-bpe} models 
compare favorably with  recently published results on MuST-C test data
(\citealt{le-etal-2021-lightweight}\footnote{28.73 on en-es, 34.98 on en-fr, 24.96 on en-it.} and \citealt{bentivogli-etal-2021-cascade}\footnote{32.93 on en-es, 28.56 on en-it.}).

Also built with the same (Transformer-based) core technology, the other systems, \textbf{\textsc{small-bpe}} and \textbf{\textsc{small-char}},
allow for apples-to-apples comparison between the different capabilities of BPE and character-level  tokenization, namely: \textit{i)} the syntactic advantage of BPE in managing several agreement phenomena \cite{sennrich-2017-how-grammatical, ataman-etal-2019-importance}, and \textit{ii)} 
the higher capability of character-level 
at generalizing morphology \cite{Belinkov2019OnTL}.
Given the morphological and syntactic nature of gender, such differences make them enticing candidates for further analysis. So far, \citet{gaido-etal-2021-split} carried out the only study interplaying the two segmententation methods and gender bias, and found that
-- in spite of lower overall performance -- character
tokenization 
results in higher production of feminine forms for ST.
%
%
%
%
%
%
%
By exploiting our new enriched resource, we intend to further test this finding and extend the analysis to 
gender agreement. 
Thus, for the sake of comparison with \citep{gaido-etal-2021-split}, we train these systems in the same (controlled) data 
conditions
i.e. on  the MuST-C corpus only.

\subsection{Evaluation method}
\label{subsec:evaluation}

We 
employ
the enriched MuST-SHE corpus
to assess generic performance and gender translation at 
several levels of granularity. 
%
%
Evaluating gender translation under natural conditions grants the advantage of inspecting diverse informative phenomena.
Concurrently, however, the intrinsic variability of natural language can defy automatic approaches based on reference translations: 
Since
language generation is an open-ended task, in our specific setting
system's outputs may not contain the exact 
gender-marked 
words annotated in MuST-SHE.
In fact, the released MuST-SHE evaluation script~\citep{gaido-etal-2020-breeding} first 
measures
dataset \textit{coverage}, i.e. the proportion of 
annotated words that are
generated by the system, and on which 
gender translation is hence measurable. Then, it calculates \textit{gender accuracy}  as the proportion of words generated in the correct gender among the measurable ones.
As a result, all the \textit{out of coverage} 
words are necessarily
left
unevaluated.

For all \textbf{word-level} gender evaluations 
(Sections
\ref{sec:results_discussion} and 
\ref{subsec:pos}), 
we compute 
accuracy as in the official MuST-SHE script and include scores based on the POS annotations.
Instead, for \textbf{chain-level} gender agreement evaluation 
(Section \ref{sec:agreement_automatic}) we 
modified the original script
to process full agreement chains instead of single words.\footnote{The scripts 
are released together with the MuST-SHE annotated extensions.}%
%
%
%
%

Finally, since we aim at gaining 
qualitative insights into
systems' behaviour, and at ensuring a sound and thorough multifaceted evaluation, we overcome the described coverage limitation of the automatic evaluation by complementing it with a manual analysis of \textit{all} the gender-marked words
and agreement chains that remained out of coverage.
This extensive manual evaluation was carried out via a systematic annotation of
systems'
outputs, performed by the same linguists that enriched MuST-SHE, who provided the appropriate knowledge of both the resource 
and the evaluation 
task.
%
Accordingly, we manage to make our study completely exhaustive by covering every gender-marked instance of MuST-SHE. Also, such additional manual evaluation serves as a proof-of-concept to ensure the validity of the employed automatic evaluation metrics.

%


\section{Word-level Evaluation}

\subsection{Overall quality and gender translation}
\label{sec:results_discussion}


%



Table \ref{tab:overall_results} presents SacreBLEU \citep{post-2018-a-call},\footnote{ \texttt{\scriptsize{BLEU+c.mixed+\#.1+s.exp+tok.13a+v.1.4.3}}} coverage, and gender accuracy scores on the MuST-SHE test sets. 
All language directions exhibit a consistent trend: 
\textsc{large-bpe}
systems
unsurprisingly
achieve by far the highest 
overall translation quality. 
Also, in line with previous analyses \cite{di-gangi-2020-target}, 
\textsc{small-bpe} models
outperform the \textsc{char} ones by $\sim$1 BLEU 
point.
The higher overall 
translation quality 
of \textsc{large-bpe} models is also reflected by the coverage scores (All-Cov), where 
they
generate the highest number of MuST-SHE gender-marked words for all language pairs. 

\begin{table}[t]
\scriptsize
\centering
\setlength{\tabcolsep}{3pt}
\centering
\begin{tabular}{cl||c|c|ccc}
\hline

\multicolumn{1}{c}{\textbf{}} & \multicolumn{1}{c||}{} & \multicolumn{1}{c|}{\textbf{BLEU}} & \multicolumn{1}{c|}{\textbf{All-Cov}} & \multicolumn{1}{c}{\textbf{All-Acc}} & \textbf{F-Acc } & \textbf{M-Acc  } \\ \hline
\hline
\textbf{}      & \textsc{small-bpe}  & 27.6  & 65.0 & 64.1 & 45.8 & 79.6   \\
\textbf{en-es} & \textsc{small-char} & 26.5  & 64.2 & 67.3 & \textbf{52.8 } & 79.6    \\
               & \textsc{large-bpe} & \textbf{34.1} & \textbf{72.0} & \textbf{69.1} & \textbf{52.8 } & \textbf{83.6} \\ 
               \hline
\textbf{}      & \textsc{small-bpe}  & 25.9  & 55.7 & 64.9 & 50.3 & 78.1   \\
\textbf{en-fr} & \textsc{small-char} & 24.2  & 55.9 & 68.5 & \textbf{57.7} & 78.2\\
               & \textsc{large-bpe} & \textbf{34.3} & \textbf{64.3} & \textbf{70.9} & 57.1 & \textbf{83.4}\\
               \hline
               \textbf{}      & \textsc{small-bpe}  & 21.0 & 53.1 & 67.7 & 52.3 & 80.3 \\
\textbf{en-it} & \textsc{small-char} & 20.7  & 52.6 & \textbf{71.6} & \textbf{57.2}     & 83.9     \\
               & \textsc{large-bpe} & \textbf{27.5}  & \textbf{59.2} & 69.1  & 52.2  & \textbf{85.4} \\
              \hline
\end{tabular}
\caption{BLEU, coverage and gender accuracy (percentage) scores  computed on MuST-SHE.}
\label{tab:overall_results}
\end{table}

By turning to overall gender accuracy (All-Acc) though,
the edge previously assessed for the bigger state-of-the-art systems ceases to be 
clear-cut. For
 en-es and en-fr, \textsc{large-bpe} systems 
outperform the concurring \textsc{small-char} by 
$\sim$2 points only --
a slim advantage compared to the 
large gap
observed on BLEU score. 
Moreover, for en-it, 
\textsc{small-char} proves the best at translating gender.
%

%
%
%

We further zoom into the comparison of gender translation for feminine (F-Acc) and masculine (M-Acc) forms,
where we can immediately assess that all ST models are
skewed toward a disproportionate production of masculine 
forms (on average, 53.1\% for F vs. 81.3\% for M).
%
%
%
%
However, focusing on \textsc{large-bpe} models, we discover that their higher global gender accuracy 
(All-Acc) is actually due to the higher generation of masculine forms, while they do not compare favorably when it comes to feminine translation.
%
In fact, in spite of achieving the lowest generic translation quality, \textsc{small-char} 
prove
on par (for en-es) or even better (for en-it and en-fr) than \textsc{large-bpe} models at handling feminine gender translation.

In light of the above, our results reiterate the importance 
of 
dedicated evaluations that, unlike holistic metrics, are able to disentangle 
gender phenomena.
As such, we 
can 
confirm that higher generic 
performance
does not entail a superior capacity of producing feminine gender. This does not only emerge, as per \citet{gaido-etal-2021-split}, in the comparison of
(small) BPE- and char-based ST models. 
Rather, even 
for stronger 
systems, we 
attest how 
profiting from a wealth of -- uncurated and synthetic \cite{parrots} -- data does not grant advantages to address gender bias.
This motivates us to 
continue our multifaceted evaluation by taking into account only small models -- 
henceforth
\textsc{char} and \textsc{bpe} -- 
that, being trained on the same  MuST-C data, allow for sound and transparent 
comparison. %

\subsection{Word classes and Parts-of-speech}
\label{subsec:pos}
\graphicspath{{imgs/}}

\begin{figure}[t]
    \centering
    \includegraphics[scale=0.1765]{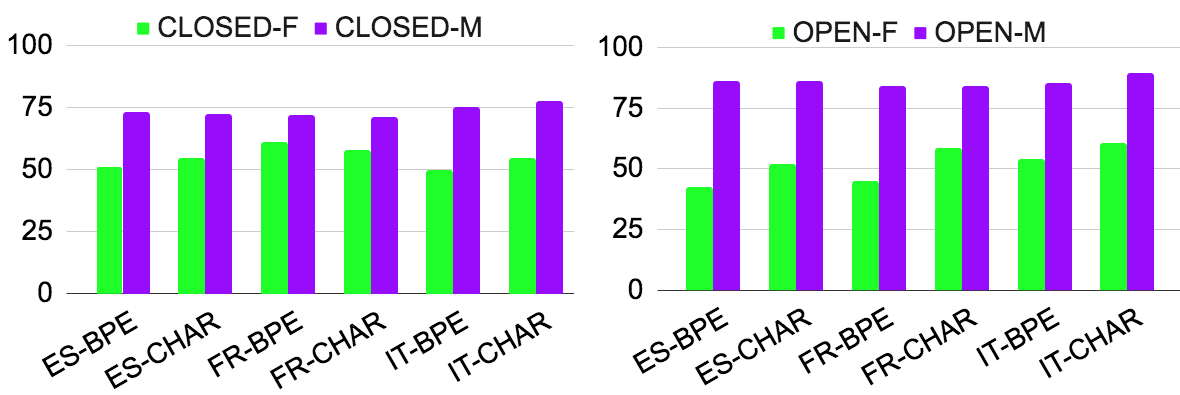}
    \captionof{figure}{
    Feminine \textit{vs}. Masculine 
    accuracy scores for closed and open class words.} 
    
        \label{fig:acc_class.PNG}
\end{figure}

\begin{table}[!]
\scriptsize
\centering
\setlength{\tabcolsep}{3pt}
\begin{tabular}{cl|cc|cc|cc}
\multicolumn{1}{l}{} &
  \multicolumn{1}{l}{} &
  \multicolumn{1}{l}{} &
  \multicolumn{1}{l}{} &
  \multicolumn{1}{l}{} &
  \multicolumn{1}{l}{} &
  \multicolumn{1}{l}{} &
  \multicolumn{1}{l}{} \\
  \hline
\multicolumn{1}{l}{} &
  \multicolumn{1}{l|}{} &
  \multicolumn{2}{c|}{\textbf{Verbs}} &
  \multicolumn{2}{c|}{\textbf{Nouns}} &
  \multicolumn{2}{c}{\textbf{Adj-des}} \\
\multicolumn{1}{l}{} &
  \multicolumn{1}{l|}{} &
  \multicolumn{1}{c}{F-Acc} &
  \multicolumn{1}{l|}{M-Acc} &
  \multicolumn{1}{c}{F-Acc} &
  \multicolumn{1}{l|}{M-Acc} &
  \multicolumn{1}{c}{F-Acc} &
  \multicolumn{1}{l}{M-Acc} \\
  \hline \hline
  \textbf{en-es} &
  \textsc{bpe} &
  44.4 &
  \textbf{93.8} &
  21.1 &
  89.0 &
  57.4 &
  \textbf{80.0} \\
 &
  \textsc{char} &
  \textbf{60.0} &
  84.2 &
  \textbf{37.4} &
  \textbf{89.7} &
  \textbf{61.2} &
  79.7\\
\hline
  \textbf{en-fr} &
  \textsc{bpe} &
  51.3 &
  \textbf{79.8} &
  16.4 &
  93.5 &
50.6 &
  78.6 \\
 &
  \textsc{char} &
  \textbf{68.4} &
  75.0 &
  \textbf{27.4} &
  \textbf{95.3} &
  \textbf{63.0} &
  \textbf{81.4} \\
  \hline
\textbf{en-it} &
  \textsc{bpe} &
  63.7 &
  83.7 &
  28.6 &
  92.2 &
  62.0 &
  76.7 \\
\textbf{} &
  \textsc{char} &
  \textbf{66.7} &
  \textbf{89.2} &
  \textbf{33.3} &
  \textbf{94.3} &
  \textbf{70.6} &
  \textbf{84.5} \\
  \hline
\end{tabular}
\caption{Feminine \textit{vs.} Masculine 
accuracy
scores per open class POS. 
}
\label{tab:POS}
\end{table}


At a finer level of granularity, we 
use 
our extension of MuST-SHE
to inspect gender bias 
across
open and closed class words.  Their coverage ranges between 74-81\% for 
function
words, 
but it shrinks 
to 44-59\% for 
content
words
(see Appendix \ref{app:class_coverage}). 
This
is expected given the limited variability and high frequency of functional items in 
language. Instead, the coverage of feminine and masculine forms is 
on par within each class 
for all systems, thus allowing us to evaluate 
gender accuracy on a comparable proportion of 
generated 
words.
%
A bird's-eye view of Figure
\ref{fig:acc_class.PNG} attests that, although masculine forms are always disproportionately produced, the gender accuracy gap is amplified on the open class words.
The consistency of such a behaviour across languages and systems  suggests 
that content words are involved to a greater extent in gender bias. 

%
%
%
%
We 
hence
analyse 
this more problematic class by looking into a breakdown of the results per POS, while for function words’ gender  
accuracy we refer to Appendix \ref{app:acuracyPOS}.
Table \ref{tab:POS} 
presents 
results for \textit{verbs}, \textit{nouns}  and \textit{descriptive adjectives}. First, in terms of 
system
capability, \textsc{char} 
still consistently emerge as the 
favorite
models 
for feminine translation. 
What we find 
notable, though,
is that  even within the same class 
we observe evident 
fluctuations, 
where nouns come forth
as the most biased POS with a huge divide between M and F accuracy 
(52--77 points).
Specifically, scores below 50\% indicate that feminine forms are generated with a probability that is below random choice, thus signalling an extremely
strong bias.



In light of this finding, we 
hypothesize
that semantic and distributional features might be a factor to interpret words’ gender 
skew.
Specifically, occupational lexicon (e.g.  lawyer, professor) makes up for most of the nouns represented in MuST-SHE ($\sim$70\%).
While such a high rate of professions in TED data is not surprising \textit{per se},\footnote{As TED talks are held by field experts,
references to education and titles are quite common \citep{mackrill2021makes}.}
it 
singles out
that professions may actually represent a category where systems largely 
rely
on spurious cues to perform gender translation, even within natural conditions that do not ambiguously prompt stereotyping. We exclude basic token frequency
by POS as a key factor to interpret our results, 
as MuST-SHE feminine nouns do not consistently appear 
as the 
POS with the lowest number of occurrences, nor 
do they have
the
lowest F:M ratio within MuST-C training data.
%
As 
discussed
in 
Section \ref{sect:ethic},
we believe that   
our 
breakdown per POS is informative inasmuch it prompts 
qualitative considerations on 
how to pursue 
gender bias mitigation in models and corpora \cite{Czarnowska2021QuantifyingSB, doughman-etal-2021-gender}.


\subsection{Manual analysis} 
\label{subsec:ooc_words}

 


We manually inspect
\textsc{char} and \textsc{bpe} system's output on the out-of-coverage (OOC) words that could not be automatically evaluated (see  ``All-Cov'' column in Table \ref{tab:overall_results})%
%
, which amount to more than 
5,000
instances. 
%
As shown 
in Table \ref{tab:Examples}, 
our analysis 
discerns between OOC words due to \textit{i)} translation \textit{errors} (Err),\footnote{
Errors range from 
misspelling to complete gibberish.}
and \textit{ii)} 
%
%
%
%
adequate {\textit{alternative} }
translations
(i.e.  meaning equivalent) for the expected gender-marked 
words.
%
%
%
%
%
Such alternatives 
comprise
instances in which  word omission is acceptable (Alt-O) \cite{baker1992coursebook}, 
and rewordings through
 synonyms or paraphrases. 
Since our focus remains on gender translation, we distinguish 
when such rewordings are generated with 
correct (Alt-C) or wrong (Alt-W) gender inflections, as well as  
neutral 
expressions devoid of gender-marking (Alt-N). 
Note that -- 
with respect to
English \cite{cao-daume-iii-2020-toward,vanmassenhove-etal-2021-neutral,sun2021they} --
overcoming the structural pervasiveness of gender specifications in grammatical gender languages 
is 
extremely challenging
\cite{neutral}, but
some rewordings 
can enable indirect neutral language (INL)%
\footnote{INL relies on generic expressions rather than gender-specific ones (e.g. \textit{service} vs. 
\textit{waiter/tress})
See 
Section
\ref{sect:ethic}.}
\cite{artemis}.

\begin{table}[!]
\centering
\setlength{\tabcolsep}{3pt}
\scriptsize
\begin{tabular}{lll}
  \toprule
  & &  \textsc{Errors}\\
  \hline
  & \textsc{src} & Robert became \textbf{fearful} and \textbf{withdrawn}. \\ 
 & \textsc{Ref}$_{it}$ & Robert divenne \textbf{timoroso} e \textbf{riservato}. \\
  & \textsc{Out}$_{it}$ & Robert diventò \textbf{timore} e \textbf{con John}. \\
  & & (\textit{Robert became fear and with John})\\
 \hline
   & & \textsc{Alternatives}\\
 \hline
 \textbf{Alt-O} & \textsc{Src} & He was \textbf{an} artist.  \\ 
     & \textsc{Ref}$_{fr}$ & C'était \textbf{un} artiste. \\
     & \textsc{Out}$_{fr}$ & C'était (\textbf{\_\_}) artiste.\\
 \hline
\textbf{Alt-C} & \textsc{src} & 
These girls [...], they are so \textbf{excited}... \\ 
& \textsc{Ref}$_{es}$ 
& 
Estas niñas [...], están \textbf{emocionad\underline{as}}...\\
& \textsc{Out}$_{es}$ & 
Estas chicas [...],
están 
\textbf{entusiasmad\underline{as}}...  \\
\cmidrule{2-3}
  \textbf{Alt-W} & \textsc{src} & Mom [...]
  \textbf{became manager...}
  \\ 
& \textsc{Ref}$_{it}$ 
& Mamma [...] 
venne 
\textbf{mess\underline{a}} a capo di... 
 \\
 & \textsc{Out}$_{it}$ & La madre 
 [...]
 diventò \textbf{cap\underline{o}} di... 
 \\
 \cmidrule{2-3}
\textbf{Alt-N} & \textsc{src} & I \textbf{felt} really good. \\ 
& \textsc{Ref}$_{fr}$  & 
Je me suis \textbf{sent\underline{i}} vraiment bien   \\
& \textsc{Out}$_{fr}$ &  Je me \textbf{\underline{sentais}} vraiment bien .\\

  
\bottomrule
\end{tabular}
\caption{Classification of OOC words.}

\label{tab:Examples}
\end{table}

\begin{figure}[t]
    \centering
    \includegraphics[scale= 0.25]{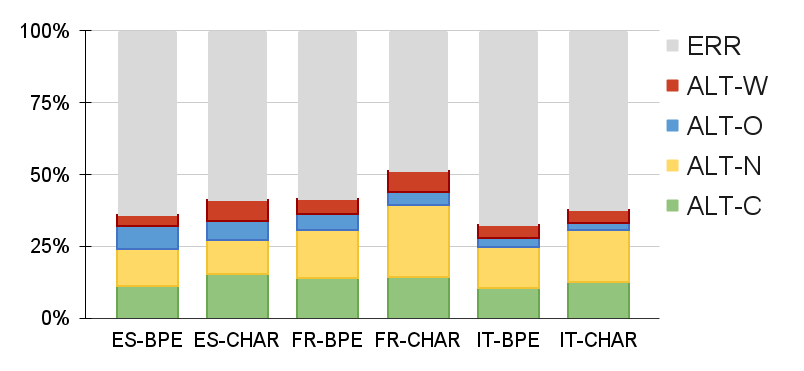}
        \captionof{figure}{Proportion of OOC words due to translation errors and alternative translations per system.
        }
\label{fig: ooc_words}
\end{figure}

The results of the analysis are shown in
Figure
\ref{fig: ooc_words}. Surprisingly, we find that \textsc{bpe} models -- in spite of their higher BLEU scores -- accumulate more translation errors than their 
\textsc{char}
counterparts.\footnote{We noticed that \textsc{char}'s lower translation quality may have to do more with fluency rather than lexical issues.} 
Conversely, \textsc{char} models generate an overall higher proportion of alternatives 
and, more importantly, 
alternatives whose gender translation
is acceptable (-N, -C).
%
This suggests that \textsc{char} output is characterized by a favourable 
\textit{adequate} variability that conveys both lexical 
meaning and gender realization better than \textsc{bpe}.
Also, note that the outcome of the manual analyses reiterates the results obtained with the automatic evaluation based on accuracy at the word-level, thus confirming its reliability. 
As a final remark, we find that 
all
systems produce a considerable amount of 
neutral alternatives in their outputs. 
To gain insight into such neutralizations, we audit on which POS they are realized. 
Accordingly, we find that neutralizations of adjectives 
and
nouns are quite limited, and concern the production of epicene synonyms (e.g. \textit{en}: happy; \textit{es-ref}: 
content\underline{o}/\underline{a}; \textit{es-out}: feliz). Verbs, instead, are largely implicated in the phenomenon, since inflectional changes in tense and aspect paradigms (e.g., present, imperfective) that do not convey gender distinctions are  feasible (see the -N 
example in Table \ref{tab:Examples}). Such range of alternatives for verbs is in fact also reflected by its lowest coverage 
among all POS (as low as {$\sim$32\%}). Finally, paraphrases 
based on
verbs also represent 
the most frequent way to neutralize other POS in the output.
%
Since such expressions 
are suitable, or even preferable, for several scenarios 
(e.g.  
to substitute masculine generics, to avoid making unlicensed gender  assumptions)
our finding encourages the creation of test sets accounting for such a third viable direction, and 
can shed light on 
systems' potential to produce INL alternatives.

\section{Gender Agreement Evaluation}

\subsection{Automatic analysis}
\label{sec:agreement_automatic}

The final step in our multifaceted analysis goes beyond the word level to inspect agreement chains in translation. 
To this aim, we  define \textit{coverage} as the proportion of generated chains matching with those annotated in MuST-SHE.  Then, the \textit{accuracy} of the generated chains accounts for 3 different cases where: 
\textit{i)} agreement is respected, and with the correct gender (C);
\textit{ii)}  agreement is respected, but with the wrong gender
(W); 
and \textit{iii)} both feminine and masculine gender inflections occur together, and thus agreement is not respected (NO).


\begin{table}[!]
\scriptsize
\centering
\setlength{\tabcolsep}{3pt}
\begin{tabular}{cc|ccc||ccc|ccc}

\multicolumn{1}{l}{} &
  \multicolumn{1}{l}{} &
  \multicolumn{1}{l}{} &
  \multicolumn{1}{l}{} &
  \multicolumn{1}{l}{} &
  \multicolumn{1}{l}{} &
  \multicolumn{1}{l}{} &
  \multicolumn{1}{l}{} \\
  \hline
\multicolumn{1}{l}{} & \multicolumn{1}{c|}{} &
\multicolumn{3}{c||}{\textbf{All}} &
\multicolumn{3}{c|}{\textbf{Feminine}} & \multicolumn{3}{c}{\textbf{Masculine}}  \\
\multicolumn{1}{l}{} & \multicolumn{1}{c|}{} &
  \multicolumn{1}{c}{C} &
  \multicolumn{1}{c}{W} &
  \multicolumn{1}{c||}{NO} &
  \multicolumn{1}{c}{C} &
  \multicolumn{1}{c}{W} &
  \multicolumn{1}{c|}{NO} &
  \multicolumn{1}{c}{C} &
  \multicolumn{1}{c}{W} &
  \multicolumn{1}{c}{NO} \\
  \hline\hline
\textbf{en-es} &
  \textit{bpe} &
  74.3	&	\textbf{24.6}	&	\textbf{1.2} & 33.9 &	\textbf{64.4} &	\textbf{1.7}	& 95.5	& \textbf{3.6}	& 0.9 \\
\textbf{} &
  \textit{char} & 
 \textbf{78.4}	&	21.0	&	0.6 & \textbf{42.4} &	57.6 &	0.0	& \textbf{96.6}	& 2.6	& 0.9 \\
  \hline
\textbf{en-fr} &
  \textit{bpe} &
  67.9	&	\textbf{31.0}	&	\textbf{1.2} & 54.1 &	\textbf{45.9} &	0.0	& 78.7	& \textbf{19.1}	& \textbf{2.1}	 \\
 &
  \textit{char} &
  \textbf{76.7}	&	22.3	&	1.0 & \textbf{57.5} &	40.0 &	\textbf{2.5}	& \textbf{88.9}	& 11.1	& 0.0 \\
  \hline
\textbf{en-it} &
  \textit{bpe} &
  71.7	&	\textbf{27.5}	&	0.7 & 47.4 &	\textbf{50.9} &	\textbf{1.8}	& 88.9	& \textbf{11.1}	& 0.0 \\
 &
  \textit{char} &
 \textbf{78.5}	&	20.0	&	\textbf{1.5} & \textbf{54.2} &	44.1 &	1.7	& \textbf{97.4}	& 1.3	& \textbf{1.3} \\
\hline
\end{tabular}
\caption{Agreement results for All chains matched in MuST-SHE, and split into Feminine and Masculine chains. 
Accuracy scores are given for agreement respected with correct gender (C), agreement respected with wrong gender (W), agreement not respected (NO).}


\label{tab:agremment-new}
\end{table}

Table \ref{tab:agremment-new} shows
accuracy scores for all MuST-SHE agreement chains (All),
also split
into feminine (F) and masculine 
(M) chains.
The overall results 
are promising: we find very few instances (literally 1 or 2) in which  ST systems produce an ungrammatical output that breaks gender agreement 
(NO).
In fact, 
both systems tend 
to be consistent with one picked gender for the whole dependency group. 
Thus, in spite of previous MT studies concluding that 
character-based segmentation 
results
in poorer syntactic capability \citep{Belinkov2019OnTL}, 
respecting concord
does not appear as an issue for 
any of our small
ST models. For the sake of comparability, however, we note that our evaluation
involves language pairs that do not widely resort to long-range dependencies; this may contribute to explaining why
\textsc{char} 
better handles correct gender agreement.\footnote{Due to space constraints we refer to Appendix \ref{app:subject-verb} for an analysis of longer-range cases of subject-verb agreement.}

Overall, agreement  translation was measured on a lower 
\textit{coverage} (30-50\%) -- presented in Appendix \ref{app:agr-coverage} --   than the word-level one 
(Section \ref{tab:overall_results}).
While this is expected given the strict requirement of generating full chains with several words, we recover such a loss by means of 
the
comprehensive manual evaluation discussed below.

\subsection{Manual analysis}
\label{subsec:agr-ooc}
Our manual inspection recovers 
a total of $\sim$1,200 OOC agreement chains from \textsc{char} and \textsc{bpe} output. 
Similarly to the approach employed for
single words
(Section \ref{subsec:ooc_words}),
we discern between OOC chains due to: \textit{i)} translation \textit{errors} 
(Err),
and \textit{ii)} 
\textit{alternative}
translations
preserving the source 
meaning.
We distinguish different types of alternatives. First, alternatives that do no exhibit a morphosyntactic agreement phenomenon to be judged, as in the case of neutral paraphrases or rewordings consisting of a single word (NO-chain).
%
Instead, when the generated alternative chain exhibits gender markings, we distinguish if the chosen gender is correct (C), wrong (W), or if the system produces  a chain that does not respect gender 
agreement because
it combines both feminine and masculine gender inflections (NO).

\begin{figure}[htp]
    \centering
    \includegraphics[scale=0.25]{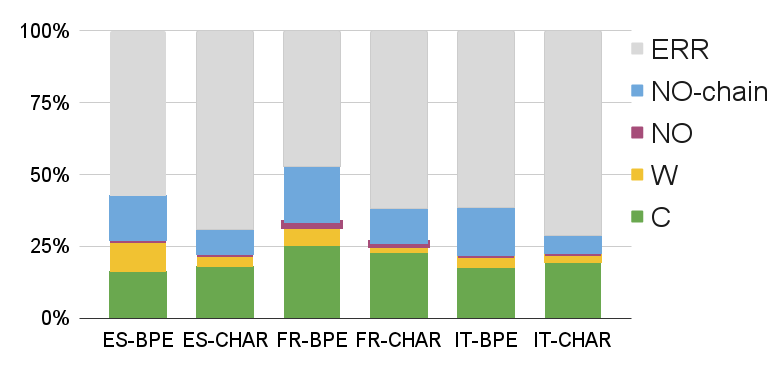}
        \captionof{figure}{{Proportion of OOC chains due to
        translation errors or alternative agreement translations per system.}}
\label{fig:agr-ooc}
\end{figure}

The outcome of such OOC chains categorization is presented in Figure \ref{fig:agr-ooc}. Interestingly, such results are only partially corroborating previous analyses. 
On the one hand, unlike the OOC words' results discussed in 
Section \ref{subsec:ooc_words},
we attest that \textsc{char} models produce the highest proportion of translation errors. Thus, it seems that \textsc{char} capability in producing adequate alternatives is confined to the single-word level, whereas it 
exhibits a
higher failure rate on longer sequences. 
On the other hand, by looking at alternative chains, \textsc{char} still 
emerges
as the best at properly translating gender agreement, 
with the highest proportion of chains with correct gender (C), and the lowest one with wrong gender (W).

Finally, again in line with 
our automatic evaluation
(Table \ref{tab:agremment-new}), we confirm that respecting agreement is not an issue 
for
our ST models: we identify only 
3 cases (2 for en-fr \textsc{bpe}, 
1
for en-fr \textsc{char}) where 
concord is broken (NO).
Given the rarity of such instances, we  are not able to draw definitive conclusions on the nature of these outliers. Nonetheless, we check the instances in which agreement was not respected 
(both in and out of coverage).
We see that cases of broken concord also concern 
extremely simple phrases, consisting of 
a noun and its modifier (e.g. en: \textit{talking to [this inventor],...because he}; fr: \textit{parler à \textbf{[cette{$_F$} inventeur{$_M$}]}}\textit{..., parce qu' il}).
However, the most common type among these outliers are 
constructions with semi-copula verbs (e.g. en: 
\textit{She... [became a vet]};
it: \textit{...E' [\textbf{diventata}{$_F$} \textbf{un}{$_M$} \textbf{veterinatrio}{$_M$}]}), which -- as discussed in 
Section \ref{subsec:categorization} --
exhibit a weaker agreement constraint. 

\section{Conclusion}
The complex system of grammatical gender languages entails several morphosyntactic implications for 
different lexical categories.
In this paper, we 
underscored 
such implications and 
explored
how different POS and grammatical agreement are 
involved
in gender bias.
To this aim, we enriched the MuST-SHE benchmark with new linguistic information, and carried out 
an extensive evaluation on the behaviour of ST models
built with different segmentation techniques and data quantities.  
On three 
language pairs 
(English-French/Italian/Spanish),
our study shows 
that, while all POS are subject to  masculine skews,
they are not impacted to the same extent. Respecting gender agreement for the translation of related words, instead,  
is not an issue for current ST models.
We also find
that ST 
generates
a considerable amount of neutral 
expressions, suitable to replace
gender-inflected ones, 
which however 
current test sets do not recognize.


Overall, our work 
reiterates
the importance of dedicated analyses that, unlike holistic metrics, can single out system's behaviour 
on
gender phenomena. Accordingly, our results are in line with previous studies showing that, in spite of lower generic performance, character-based segmentation exhibits a better capability at handling
feminine translation at different levels of granularity.  
As our MuST-SHE extension is available for both ST and MT,  
we invite MT studies to start from our discoveries and resource. 

\section*{Acknowledgments}

%
%
%
%
We would like to thank the 2021 Summer Internship students at FBK for their contribution: Francesco Fernicola, Sara Giuliani, Lorena Rocio Mart{\'i}n, Silvia Alma Piazzolla, M{\'e}lanie Prati, Jana Waldmann. This work was made possible thanks to their extensive annotation work  
and 
active participation in fruitful discussions.


\section{Impact statement}
\label{sect:ethic}

In this paper, we evaluate whether and to what extent ST models exhibit biased behaviors by systematically and disproportionately favoring masculine forms in translation. Such a behavior is problematic inasmuch it leads to under-representational harms by reducing feminine visibility \cite{blodgett-etal-2020-language, savoldi2021gender}.

\noindent{\textbf{Broader impact.}} While the focus of this work is on the analysis 
itself, our insights 
prompt broader considerations. Specifically, our investigation on the relation between data
size/segmentation technique and gender bias provides initial cues on which models and components to audit and
implement
toward the goal of reducing gender bias. This, in particular, may be informative to define the path for emerging 
direct ST technologies.
Also, our results disaggregated by POS 
invite
reflections on how 
to intend and mitigate bias by means 
of interventions
on the training data. In fact, while it is known that the MuST-C corpus \cite{MuST-Cjournal} used for training comprises a 
majority of masculine speakers,\footnote{\url{https://ict.fbk.eu/must-speakers/}} the fact that certain lexical categories are more biased than others 
suggests
that, on top of more coarse-grained quantitative 
attempts at gender balancing \cite{costa-jussa-de-jorge-2020-fine}, data curation ought to account for more sensitive, nuanced, and 
qualitative
asymmetries. These also imply \textit{how}, rather than only \textit{how often}, 
gender groups
are represented 
\cite{wagner2015s, devinney-etal-2020-semi}. Also, while nouns come forth as the most problematic POS, current practices of data augmentation based on a pre-defined occupational lexicon may address stereotyping \cite{saunders-byrne-2020-reducing}, but do not increase the production 
of
other nonetheless skewed 
lexical categories.
Overall, our enriched resource\footnote{It will be released under the same CC BY NC ND 4.0 International  license as MuST-SHE.} can be 
useful 
to monitor the validity of different technical interventions. 

\noindent{\textbf{Ethic statement.}} 
The use of gender as a variable \cite{larson-2017gender}
warrants some ethical reflections. 

Our evaluation on the MuST-SHE benchmark exclusively accounts for linguistic gender expressions. As reported in 
MuST-SHE data statement \cite{bender-friedman-2018-data},\footnote{\url{https://ict.fbk.eu/must-she/}} also for the subset of sentences that contain first-person references\footnote{Category 1 in the corpus.} (e.g. \textit{I’m a student}), speakers’ gender information is manually annotated based on the personal pronouns found in their publicly available personal TED profile, and used to check that the indicated (English) linguistic gender forms are rendered in the gold standard translations. 

While our experiments are 
limited to the binary linguistic forms represented in the used data, to the best of our knowledge, ST natural language corpora going beyond binarism do not yet exist.\footnote{\citet{saunders-etal-2020-neural} enriched WinoMT to account for non-binary language. While it is only available for MT, such annotations consist of placeholders for neutrality rather than actual non-binary expressions. 
} This is also due to the fact that unlike English – which finds itself for several cultural and linguistic reasons as a leader of change toward inclusive forms \cite{ackerman2019syntactic} -- Direct Non-binary Language 
based on neomorphemes \cite{shroy2016innovations,papadopoulos2019morphological,knisely2020franccais} is non-trivial to fully implement in grammatical gender languages \cite{Hellinger, gabriel2018neutralising} and still 
object of 
experimentation 
\cite{effequ,artemistrans}. However, our manual evaluation expands to the possibility of INL
strategies that could be detected in system’s output. We underscore that such strategies are recommended and fruitful to avoid the gendering of referents, but are to be considered as concurring to -- rather than replacements of -- emerging linguistic innovations \cite{artemis}. 

Lastly, we signal that direct ST models may leverage speakers’ vocal characteristics as a gender cue to infer gender translation. Although the potential risks of 
such condition do not emerge and are not addressed in our 
setting (focused on POS and agreement features as a variable),
%
we endorse the point made by \citet{gaido-etal-2020-breeding}. 
Namely, direct ST systems leveraging speaker’s vocal biometric features as a gender cue can entail real-world dangers, like the categorization of individuals by means of biological essentialist frameworks \cite{zimmantransgender}. This can reduce gender to stereotypical expectations about how masculine or feminine voices should sound, and can be especially harmful to transgender individuals, as it can lead to misgendering \cite{stryker2008transgender} and invalidation. 
%
Note that we experimented with unmodified models for the sake of hypothesis testing without adding variability, but real-world deployment of ST technologies must account for the potential harms arising form the use of direct ST technologies \textit{as is}.

\bibliography{custom}
\bibliographystyle{acl_natbib}


\appendix

\section{MuST-SHE Manual Annotation}
\label{app:annotations}
POS and agreement chains annotations were carried out on MuST-SHE reference translations. 
To ensure precision, the two layers of linguistic information have been added \textit{i)} as two separate annotation processes; \textit{ii)} following strict and comprehensive guidelines. 

A first version of the guidelines was written by one of the authors -- who is 
an expert linguist -- based on a preliminary manual analysis of a MuST-SHE sample. Successively, such guidelines have been refined and improved by means of discussions with the annotators, who had carried out a pilot annotation round to get acquainted with MuST-SHE language data. 
The final version of the annotation guidelines is included in the resource release (\url{ict.fbk.eu/must-she}) and is also retrievable at: \url{https://bit.ly/3CdU50s}.
%
%


The 6 annotators were all interning students undergoing a MA degree in Linguistics/Translation Studies, who were selected among other 120 candidates. 
We ensured that at least one annotator per language was a native speaker, whereas the second one had at least a C1 proficiency level of the assigned language.
Since the annotations were carried out in the course of this more extensive curricular internship, there was no task-associated compensation. 





\section{ST Models}
\label{app:models}
In this section we describe in detail the ST models created for our study, whose source code is publicly released at: \url{https://github.com/mgaido91/FBK-fairseq-ST/tree/acl_2021}.


\subsection{Architecture}
The architecture of our ST models is composed of two strided 2D convolutional layers with 64 3x3 kernels, followed by a Transformer \citep{vaswani2017attention} with 11 encoder layers and 4 decoder layers. The two 2D convolutions reduce the length of the input sequence by a factor of 4, allowing the processing by the Transformer encoder layers that have a quadratic memory complexity with respect to the input length (because of the self-attention mechanism). The weights of the encoder self-attention matrices are biased to be close to 0 for elements far from the matrix diagonal (i.e. for elements that are far from the considered vector) with a logarithmic distance penalty 
\cite{di-gangi-etal-2019-enhancing}.
In both encoder and decoder Transformer layers, we use 8 attention heads, 512 embedding features, 
and 2048 features for the feed-forward inner-layers. The resulting number of parameters is 60M for BPE models and 52M for character-based models.

\subsection{Data}

Since the amount of ST data is limited (i.e. MuST-C -- \citealt{MuST-Cjournal} -- and Europarl-ST -- \citealt{europarlst}),
knowledge transfer from the ASR and MT tasks is leveraged 
by
respectively  
initializing the
ST encoder with ASR pretrained weights \citep{weiss2017sequence,bansal-etal-2019-pre} and 
by distilling knowledge from a strong MT teacher \citep{liu2019endtoend}.
The ASR model used for the pretraining has been trained on Librispeech \citep{librispeech}, Mozilla Common Voice,\footnote{\url{https://voice.mozilla.org/}} TEDLIUM-v3 \citep{Hernandez_2018}, and the utterance-transcript pairs of the ST corpora and of How2 \cite{sanabria18how2}.
The teacher MT models, instead, are trained on a subsample of the Opus \citep{opus} repository, filtered using the cleaning pipeline of ModernMT.\footnote{\url{https://github.com/modernmt/modernmt}}

SpecAugment is applied to the source audio with probability 0.5 by masking two bands on the frequency axis (with 13 as maximum mask length) and two on the time axis (with 20 as maximum mask length).
Time stretch~\cite{nguyen2019improving} alters the input utterance with probability of 0.3 and stretching factor sampled uniformly for each utterance between 0.8 and 1.25 is also used to alter the input audio. 
The target text was tokenized with Moses.%
\footnote{\url{https://github.com/moses-smt/mosesdecoder}}
We normalized audio per-speaker and extracted 40 features with 25ms windows sliding by 10ms with XNMT\footnote{\url{https://github.com/neulab/xnmt}} \cite{XNMT}.

The \textsc{large-bpe} model is trained on all the available (ST and distilled) data for a total of $\sim$1.25M pairs, while the \textsc{small-bpe} and \textsc{small-char} are trained only on the MuST-C data for a total of 250-275k pairs.
The encoder pretraining is used for all the models. The \textsc{small-*} models are initialized with the weights of an ASR trained only on the \textit{(audio, transcript)} pairs of MuST-C, while the \textsc{large-bpe} is initialized with an ASR trained on all the available data.

For the small and large models leveraging BPE, we employed 8k merge rules, while we used a set of 250-400 characters for the  \textsc{small-char} model. The resulting vocabulary sizes are reported in Table~\ref{tab:dictsize}.

\begin{table}[h]
\centering
\small
\begin{tabular}{l|ccc}
& \textbf{en-es} & \textbf{en-fr} & \textbf{en-it} \\\hline
Large-BPE     & 11,940 & 12,136 & 11,152 \\
Small-Char   &  464 &  304  & 256 \\
Small-BPE    &  8,120 &  8,048 & 8,064 \\
\end{tabular}
\caption{Resulting 
vocabulary
sizes.} 
  \label{tab:dictsize}
\end{table}

\subsection{Training procedure}

The models are trained using label smoothed cross-entropy \cite{szegedy2016rethinking} -- the smoothing factor is 0.1 -- with Adam using $\beta_1$=0.9, $\beta_2$=0.98 and the learning rate is linearly increased during the warm-up phase (4k iterations) up to the maximum value $5\times 10^{-3}$, followed by decay with inverse square root policy. The dropout is set to 0.2. Each mini-batch consists of 8 samples, we set the update frequency to 8, and we train on 4 GPUs, so a batch contains 256 samples.

The
\textsc{large-bpe} training is performed
in three consecutive steps. First, we train on synthetic data obtained by automatically translating the ASR corpora transcript with our MT model \citep{jia2018leveraging}. Second, we fine-tune on the ST corpora. In both these training phases the model is optimised to learn the output distributions of the MT teacher (via word-level knowledge distillation). Lastly, the model is fine-tuned on the ST corpora using label-smoothed cross entropy. 
Trainings are stopped after 5 epochs without improvements on the validation loss and we average 5 checkpoints around the best on the validation set. In the rich-data condition case, as we did not see benefits by the average of the checkpoints, we used the best checkpoint instead.
As a validation set we rely on the MuST-C gender-balanced dev set \cite{gaido-etal-2020-breeding}.

Our code is built on the Fairseq library \cite{ott2019fairseq} and trainings are performed on 4 K80 GPUs, lasted 4 days for the MuST-C-only trainings and 12 days for the rich-data models.



\section{Word-level Evaluation}

\subsection{Coverage per open and closed class words}
\label{app:class_coverage}

\begin{figure}[htp]
    \centering
  \includegraphics[width=\linewidth]{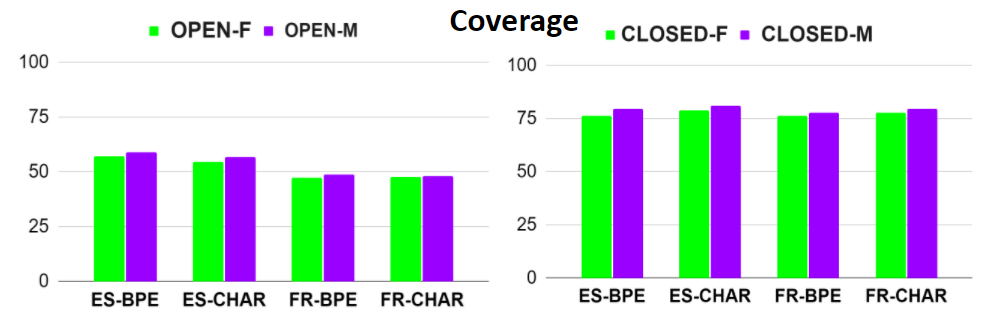}
    \caption{Feminine \textit{vs} Masculine 
    coverage scores per open and closed class words. }
    \label{fig:clas-coverage-new}
\end{figure}

As we can see in Figure \ref{fig:clas-coverage-new}, function words have a higher coverage than content words.
 This is expected given the limited variability and high frequency of functional items in language. Instead, the coverage of feminine and masculine forms is on par within each class for all systems, thus allowing us to evaluate gender accuracy on a comparable proportion of generated words.


\subsection{Gender accuracy per closed class POS}
\label{app:acuracyPOS}
\begin{table}[h]
\scriptsize
\centering
\setlength{\tabcolsep}{3pt}
\begin{tabular}{cc|cc|cc|cc}
\hline
\multicolumn{1}{l}{} &
  \multicolumn{1}{l|}{} &
  \multicolumn{2}{c|}{\textbf{Art}} &
  \multicolumn{2}{c|}{\textbf{Pronoun}} &
  \multicolumn{2}{c}{\textbf{Adj-det}} \\
\multicolumn{1}{l}{} &
  \multicolumn{1}{l|}{} &
  \multicolumn{1}{c}{F-Acc} &
  \multicolumn{1}{l|}{M-Acc} &
  \multicolumn{1}{c}{F-Acc} &
  \multicolumn{1}{l|}{M-Acc} &
  \multicolumn{1}{c}{F-Acc} &
  \multicolumn{1}{l}{M-Acc} \\
  \hline\hline
\textbf{en-es} &
  \textit{bpe} &
  51.35 &
  \textbf{70.0} &
  \textbf{52.0} &
  84.9 &
  49.1 &
  86.1 \\
 &
  \textit{char} &
  \textbf{53.5} &
  68.4 &
  51.7 &
  \textbf{85.7} &
  \textbf{59.3} &
  \textbf{91.2} \\
  \hline
  \textbf{en-fr} &
  \textit{bpe} &
  \textbf{52.0} &
  \textbf{69.2} &
  \textbf{65.5} &
  \textbf{78.3} &
  \textbf{82.9} &
  \textbf{79.5} \\
 &
  \textit{char} &
  50.8 &
  68.6 &
  54.2 &
  77.3 &
  79.1 &
  78.6 \\
  \hline
  \textbf{en-it} &
  \textit{bpe} &
  47.2 &
  74.6 &
  \textbf{75.0} &
  71.4 &
  50.9 &
  81.8 \\
\textbf{} &
  \textit{char} &
  \textbf{52.2} &
  \textbf{76.8} &
  52.9 &
  \textbf{77.8} &
  \textbf{61.8} &
  \textbf{83.3} \\
  \hline
\multicolumn{1}{l}{} &
  \multicolumn{1}{l}{} &
  \multicolumn{1}{l}{} &
  \multicolumn{1}{l}{} &
  \multicolumn{1}{l}{} &
  \multicolumn{1}{l}{} &
  \multicolumn{1}{l}{} &
  \multicolumn{1}{l}{} \\
\end{tabular}
\caption{Feminine \textit{vs.} Masculine \bs{percentage} accuracy scores per closed class POS.}
\label{tab:closepos}
\end{table}

As we can see in Table \ref{tab:closepos}, \textsc{char}'s otherwise attested advantage over \textsc{BPE} is not consistent for function words, where we find variations across POS and languages.  
Such variations may be due to the fairly restricted amount of MuST-SHE \textit{pronouns} and \textit{limiting adjectives} (Adj-det) on which accuracy can be computed in MuST-SHE (see Table \ref{tab:annotation}), 
which make very fine-grained evaluations particularly unstable. Additionally -- since the present POS evaluation still remains at the word level -- we are not able to ponder whether gender translations for modifiers (i.e. articles, determiners) is to some extent constrained by the content words they refer to. We intend to explore such hypothesis in future work by intersecting POS and agreement annotations.


\section{Agreement Evaluation}
\label{app:agreement}

\subsection{Agreement coverage}
\label{app:agr-coverage}

\begin{figure}[htp]
    \centering
  \includegraphics[scale=0.40]{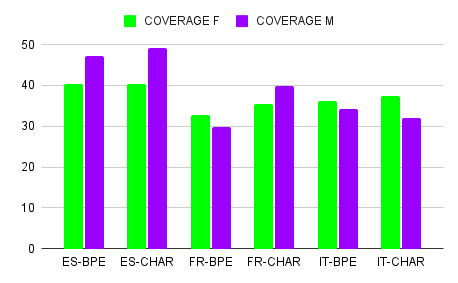}
    \caption{Feminine \textit{vs} Masculine 
    coverage scores for chains.}
    \label{fig:agr-coverage-new}
\end{figure}

Figure \ref{fig:agr-coverage-new} shows coverage of fully generated agreement chains split into feminine (F) and masculine (M) forms. Although we attest notable variations across languages and gender forms, overall masculine and feminine chains are both produced at comparable rate. 







\begin{table}[!]
\centering
\setlength{\tabcolsep}{3pt}
\scriptsize
\begin{tabular}{lll}
  \toprule
 & \textsc{src} & \textbf{A} young \textbf{scientist} that I was working with ..., Rob, \textbf{was}.. \\ 
 & \textsc{Ref}$_{it}$ & \textbf [\underline{\textbf{Un}} giovane \textbf{\underline{scienziato}}] con cui lavoravo ..., Rob, è \textbf{stato}..\\

 \toprule
\end{tabular}
\caption{Example of subject-verb agreement in MuST-SHE.}
\label{tab:sub-verb}
\end{table}

\subsection{Manual analysis of subject-verb agreement}
\label{app:subject-verb}


Considering long-range dependencies that go beyond the phrase level,
a gender relevant co-variation is also that of \textit{subject-verb} agreement, as the one shown in Table \ref{tab:sub-verb} (see also footnote 1).
To account for such longer spans, we considered all MuST-SHE sentences where both \textit{i)} a word (or chain) functioning as a subject, and \textit{ii)} its referring verb or predicative complement are annotated as gender-marked words in the references.  We identified 55 sentences for en-es, 54 for en-fr and 41 for en-it, and we manually analyzed all the corresponding systems' outputs.





We found that, even in the case of dependencies within a longer range, systems largely respect agreement in translation and consistently pick the same gender form for all co-related words. In fact, we identified only 4 cases where concord is broken: 1 case  each for \textsc{bpe} and \textsc{char} for en-es and en-it, and none for en-fr. 
%
%
Among these 4 cases, we found the above discussed weaker gender-enforcing structures (see the description of (semi-)copula verbs and their predicative complements  in  
Section
\ref{subsec:agr-ooc}), and we also
detected what resembles \textit{agreement attraction errors} \cite{linzen2016assessing}. %
Namely, the model does not produce a verb or complement in agreement with its actual (but distant) subject, as other words intervene in the sentence and agreement is conditioned by the verb/complement's preceding word. As a result, subject-verb agreement is not respected. 
The following (long) sentence is an example of such an attraction error, where the complement \textit{desperate} is inflected in the same masculine and plural form as its just preceding noun rather than the chain functioning as subject (\textit{the nurse}).

\bigskip


\noindent\fbox{\begin{minipage}
{19em}
\footnotesize{(\textit{en-src}) I watched in horror heartbreaking footage of \textbf{the head nurse}, Malak, in the aftermath of the bombing, grabbing premature babies out of their \underline{incubators}, \textbf{desperate} to get them to safety, before she broke down in tears.

(\textit{es-\textsc{char}}: Vi una imagen horrible desgarradora de \textbf{l\underline{a} enfermer\underline{a}} (F., sing.) mi laguna, en los ratones después del bombardeo, agarrando a los bebés permaturos fuera de sus \underline{incubadores} (M., pl.) \textbf{desesperados}(M., pl.) por hacerlos...}
\end{minipage}}

\bigskip

Such kind of agreement issues have more to do with  overall syntactic capacity of ST models, rather than being implicated  with gender bias. We can thus conclude that, even taking into account longer dependencies, agreement still does not emerge as an issue entrenched with gender bias.

 \end{document}